\documentclass[a4paper,fleqn]{cas-dc}

\usepackage{lipsum}

\usepackage[numbers]{natbib} 
\usepackage{graphicx}
\usepackage{todonotes}
\usepackage{xcolor}
\usepackage{flushend}

\begin{document}
\sloppy
\let\WriteBookmarks\relax
\def\floatpagepagefraction{1}
\def\textpagefraction{.001}
\shortauthors{Erhan et al.}
\shorttitle{Smart Anomaly Detection in Sensor Systems: {A Multi-Perspective Review}}

\title [mode = title]{Smart Anomaly Detection in Sensor Systems: \\ {A Multi-Perspective Review} }   

\author{{L. Erhan, M. Ndubuaku, M. Di Mauro, W. Song, M. Chen, G. Fortino, O. Bagdasar,  A. Liotta}}

\begin{abstract}[S U M M A R Y]
Anomaly detection is concerned with identifying data patterns that deviate remarkably from the expected behaviour. This is an important research problem, due to its broad set of application domains, from data analysis to e-health, cybersecurity, predictive maintenance, fault prevention, and industrial automation. Herein, we review state-of-the-art methods that may be employed to detect anomalies in the specific area of sensor systems, which poses hard challenges in terms of information fusion, data volumes, data speed, and network/energy efficiency, to mention but the most pressing ones. In this context, anomaly detection is a particularly hard problem, given the need to find computing-energy-accuracy trade-offs in a constrained environment. We taxonomize methods ranging from conventional techniques (statistical methods, time-series analysis, signal processing, etc.) to data-driven techniques (supervised learning, reinforcement learning, deep learning, etc.). We also look at the impact that different architectural environments (Cloud, Fog, Edge) can have on the sensors ecosystem. The review points to the most promising intelligent-sensing methods, and pinpoints a set of interesting open issues and challenges.     

\end{abstract}

\begin{keywords}
Anomaly detection \sep
Machine learning \sep
Sensor systems \sep
Internet of Things \sep
Intelligent sensing \sep
\end{keywords}

\maketitle

\section{Introduction}\label{introduction}

Thanks to the hyper-connectivity of the Internet of Things (IoT), electronic sensors and sensor systems  have become major generators of data, currently reaching yearly rates on the zettabyte (i.e., a trillion gigabyte) scale \cite{idc_2025}. This ever-increasing amount of data has reached the big-data sphere \cite{big_data_iot_survey}, not only for their sheer volume but, especially, in terms of variety, velocity, veracity, and variability. Thus, relying merely on Cloud-assisted computing for the analysis of sensor data would pose too much of a burden on the network. 
In this context, a vast body of researchers has been investigating a range of methods for detecting anomalies in sensor systems, a critical building block in IoT networks and systems. 

Our aim is to review methods, ranging from conventional techniques (statistical methods, time-series analysis, signal processing, etc.) to data-driven techniques (supervised learning, reinforcement learning, deep learning, etc.). Moreover, we  look at the impact that different architectural deployments (Cloud, Fog, Edge) could have on the sensors world. The review takes into account the most interesting smart sensing methods, and identifies a set of appealing open issues and challenges.     

Anomaly detection \cite{AD_survey_Chandola} is a much broader problem, going well beyond the sensor systems that we scrutinize herein, and dating back many years in the research panorama. It pertains a vast number of application domains, each one with its peculiarity and constraints. Prominent examples, beyond the general fields of data analysis and artificial intelligence, are cybersecurity, predictive maintenance, fault prevention, automation, and e-health, to mention but a few.

More generally, anomaly detection is concerned with identifying data patterns that  deviate remarkably from the expected behaviour. This is crucial in the process of finding out important information about the system functioning, detecting abnormalities that are often rare or difficult to model or, otherwise, to predict \cite{DBLP:books/sp/Aggarwal2013}.  
A timely identification of anomalies is crucial to tackling a number of underlying problems that, if undetected, may lead to costly consequences. Examples are: spotting stolen credit cards; preventing systems failure; or anticipating cancer occurrence.

Anomaly detection has conventionally been tackled from the statistical viewpoint. The prominence of machine learning (ML) has, however, opened new possibilities for the detection of outliers, particularly thanks to the availability of vast amounts of data to train sophisticated learning models. This is an attractive proposition, particularly in domains such as the IoT, whereby new data patterns make it difficult to use static models \cite{atzori_internet_2010,Fortino_book}.

Sensor systems play an important role in modern interconnected digital infrastructures, for instance in environmental monitoring, smart cities, factory automation, autonomous transportation, and intelligent buildings. Typically, multiple sensors, including heterogeneous ones, come together to form a system, whereby neighbouring devices can communicate with each other and transfer data to a Cloud infrastructure for further elaboration \cite{fortino_iot}. 
Given the many types of sensing devices,  the type of data that flows through the system can vary greatly, in either format, shape (in time and space) and semantics. That is why the process of separating out normal from abnormal sensed data is a particularly challenging one.

\begin{figure*}[]
	\centering
	\includegraphics[scale=0.2]{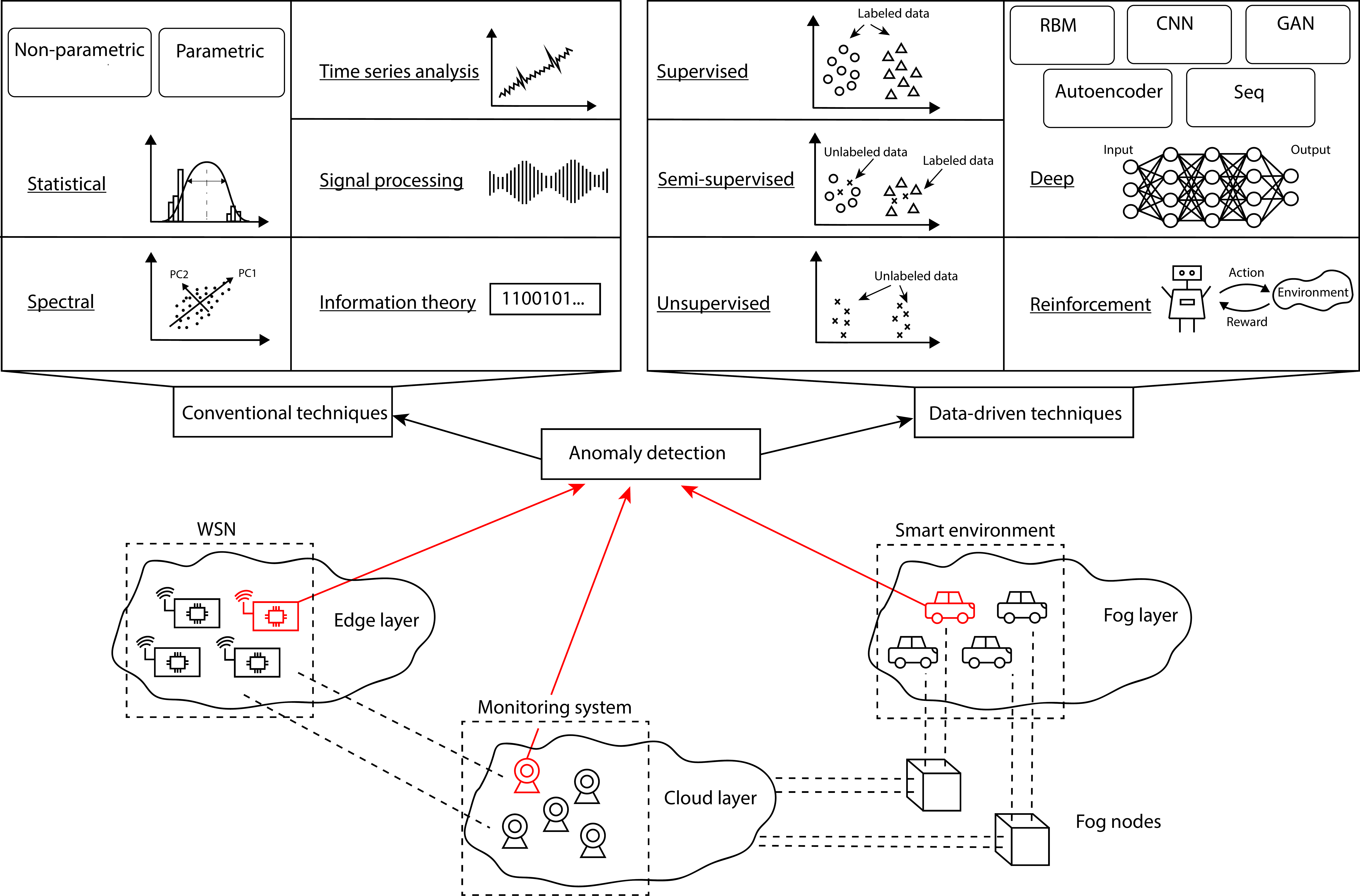}
	\caption{High perspective on sensor networks with focus on: anomaly detection techniques (conventional vs. data-driven), and architectural models (Cloud, Fog, Edge).}
	\label{fig:feedback_impl}
\end{figure*} 

In the context of IoT applications, sensors are the real source of big data, which suggests that anomaly detection at the Edge could be a powerful vehicle to address the inevitable data communication bottlenecks. An example is offered by an analysis by Cisco \cite{cisco_2019}, which estimated that a smart city with 1 million inhabitants generates daily data rates at the tunes of 180 Petabytes. The real challenge is to efficiently extract the useful information, out of this deluge of mostly useless raw data. The aim would be to considerably limit the transmission (and subsequent storage and processing) of sensor data from the network Edge to the data centres.

To this end, machine learning methods are being investigated as a promising way to automate the process of collecting data (at the sensors) for the purpose of analysis, first locally and, then, in the Cloud. This fits with the definition of machine learning (ML) by Murphy \cite{Murphy_ML}, as a set of methods that can automatically detect patterns in data, and then use the uncovered patters to predict future data, or to perform other kinds of decision making under uncertainty. This notwithstanding, such methods still need to be refined and improved to tackle the vast amount of data produced by sensor systems \cite{surveyML_big_data_processing_Qiu}.

Another peculiarity of sensor data, compared to other data-intensive systems, is the tendency to shift from offline data processes to real-time (or near real-time) requirements, whereby a timely elaboration of the raw data into usable information becomes a crucial necessity. Yet, not only industrial IoT processes but also other mission-critical domains (such as smart cities or disaster recovery applications) rely on anomaly detection (along with other data processing functionality). This criticality has led to a range of studies on how to best distribute data processes between the system's Edge (sensor nodes and aggregators) and the Cloud infrastructure. As a response, a broad range of studies have looked at the problem of shifting some processes from the data-intensive Cloud facilities to Edge devices and sensors, which are severely constrained in terms of computing and energy budget \cite{shafique_overview_roadmap_MLiot_2018}.

In fact, in this paper we argue that this architectural shift from Cloud-centred to Cloud-assisted processing (as illustrated in Sect.\ref{sec:arch}) pertains not only to conventional anomaly detection models (Sect.\ref{conv_techniques}) but also to the emerging data-driven (and even machine learning) algorithms (Sect.\ref{data-driven-techniques}).    
This trend has also been documented in  \cite{yu_survey_edgeforIoT_2018}, whereby the focus was on comparing IoT deployments on different architectural models, based on Cloud and Edge computing, respectively. Similarly, authors in \cite{pace_intelligence_edge_complexnetworks_2019} suggest that anomaly detection at the Edge of the network can act as an important building block for Edge intelligence, including pre-processing, filtering and data compression tasks. In turn, Edge computing allows for scaling up IoT systems and communications.

An interesting body of surveys have appeared in the literature, touching aspects that are relevant or complimentary to this paper. It is worth mentioning a selection of papers that have provided a snapshot of the state-of-the-art at different points in time, particularly when machine learning was still not sufficiently mature for Edge computing.  
A 2004 survey by Hodge and Austin, provides a general overview of methodologies for outlier detection, both statistical and machine learning based \cite{hodge_outlier_detection_methodologies_2004}. A broad overview of anomaly techniques spanning diverse research areas and application domains is given in \cite{AD_survey_Chandola}. Neither works are specifically targeting sensor systems, which is our focus. Also, important recent developments in lightweight machine learning methods, in connection with Fog and Edge computing, could have not been captured.   

Useful, background information relevant to this paper can be found in 
\cite{AD_evolving_data_Salehi}, which presents anomaly detection techniques for evolving data, namely data streams and evolving graphs. Techniques for discrete sequences have been introduced in \cite{chandola_ad_discrete_sequences_2012}. A perspective on intrusion detection in IoT systems is given in \cite{behniafar_ad_iot_intrusion_survey_2018}.

An interesting set of review papers by Markou and Singh provide a comprehensive background on novelty detection techniques, looking specifically at neural networks \cite{markou_novelty_nn_2003} and statistical approaches \cite{markou_novelty_statistics_2003}, with an updated review in \cite{pimentel_review_ND_2014}. 
Those papers focus on presenting the reader with a wide range of techniques and in depth explanations, alongside their strengths and weaknesses. While their emphasis is on computational -- intensive methods, we focus on the sensor systems issues and constraints, looking at advancements in light-weight machine learning and architectural paradigms that allow for  anomaly detection closer to network Edge.

More closely related to our paper is the work by Xie et al.
\cite{xie_anomaly_wsn2011}, offering a review of anomaly detection techniques, specifically for Wireless Sensor Networks (WSNs). The authors introduce key design principles to be considered for anomaly detection in WSNs, including WSNs security and node failure topics. Interestingly, they introduce a number of exemplary cases for each category and research strand, specifically for WSNs, along with two types of WSN architectures, namely flat and hierarchical ones. In our review, we consider architectural approaches and computational models that had not reached maturity back in 2011, looking at: 1) the broader sensor systems ecosystem; 2) Fog, Cloud, Edge, and distributed computing for sensor systems; and 3) lightweight ML for constrained devices.


With our paper we are aiming to provide an up-to-date snapshot of anomaly detection techniques, with a specific focus on methods suitable for sensor systems and Cloud-assisted sensing. This involves both data-intensive methods, suited for Cloud computing, and lightweight methods, aimed Edge and in-node computing. The context of our review is exemplified in Fig.\ref{fig:feedback_impl}, where we depict the software architectural elements (bottom) and a taxonomy of the methods considered for sensor systems (top).

The paper is organized as follows. Sect.\ref{anomalies}  provides an introduction to anomaly detection, and places it in the particular context of sensor systems. Our taxonomy considers conventional, more established techniques for anomaly detection in Sect.\ref{conv_techniques}. We then consider data-driven methods for anomaly detection (Sect.\ref{data-driven-techniques}), looking closely at the options that machine learning offers for sensor systems. We then switch to an architectural viewpoint (Sect.\ref{sec:arch}), discussing options to deploy anomaly detection processes in the Cloud, in the Fog or at the Edge. We finalize with a glimpse of what the future may hold for the field (Sect.\ref{open-issues}), discussing interesting research issues and challenges. 

This is a particularly prolific investigation domain, where both the software architectures and the computations methods are evolving, and there is a trend to push intelligence toward the Edge, closely to where the data is actually generated. We discuss the miniaturization and acceleration of data-intensive methods, energy efficiency, hierarchical learning models, and data heterogeneity and fusion.

\section{About anomalies}\label{anomalies}

This section provides a general introduction to anomaly detection, for the benefit of the non-specialist reader. In the second part, we put anomalies in the particular context of sensor systems.   

\subsection{The concept of anomaly}\label{anomalies_general}

Anomalies are also known as outliers, abnormalities, deviants. Hawkins \cite{hawkins_identification_outliers} defines an outlier as \textit{"an observation, which deviates so much from other observations as to arouse suspicions that it was generated by a different mechanism."} 
An important thing to consider in anomaly detection is the type of the anomaly. Anomalies could be classified in the following categories \cite{AD_survey_Chandola}:
\begin{itemize}
    \item point anomalies: when an individual data point is different from the rest of the data;
    \item contextual anomalies: when a data instance is anomalous in a specific context only, meaning that in all other situations it would be perceived as normal;
    \item  collective anomalies: when a group of related data points is anomalous compared to the dataset; the individual data points could represent normality, while it is their actual sequence that represents an anomaly.
\end{itemize}

Another important aspect to consider is the type of input data, as this requires certain techniques and poses a number of challenges. Some details to consider are: dimensionality (unidimensional, or multidimensional); number of attributes (univariate, or multivariate); and any existing relationship between the data instances. It is common for the data points to be related with each other, as it is the case for sequence data, spatial data, and graph data \cite{AD_survey_Chandola}. For sequence data, also known as temporal data, the data points are linearly ordered, representing an ordered series of events. Common examples include time series (both discrete and continuous), genome sequences, and so on. Spatial data consist of points that are related in space. Data can be both spatial and temporal, as it is the case for images and video, and it is referred to as spatiotemporal. 
Other types of data which pose difficulties for anomaly detection include:
\begin{itemize}
    \item Evolving data: the environment which one monitors to detect anomalies can be non-stationary, meaning that the data changes in time, as the characteristics of the system dynamically change. As a result, the models and algorithms need to be adaptive and to account for the changes which occurred. Classic stationary algorithms do not work as the underlying data distribution changes constantly.
    \item Streaming data: it is a sequence of data points that is mainly generated by real-time data sources. It is characterized by a continuous flow and a high number of data instances over a short period of time. It poses multiple technical challenges such as storing and processing the data on the go or online. Furthermore, the data can also evolve and change throughout time.
    \item Correlated data: in certain situations, when multiple data sources are used to monitor one particular system, it is often likely that the generated data streams are correlated. In these cases, for detecting anomalies, it is worth analyzing also the combination of  the data streams as certain data instances may not be anomalous by themselves, but they may indicate an anomaly when looked at all-together. The data instances that can be related to each other can also be part of the same stream, as it is the case with sequence data. In the latter case, in order to detect anomalies, one needs to analyze a sequence of data instances of a certain length.
    \item Heterogeneous data: in IoT systems, it is often the case that the collected data is heterogeneous. Generally speaking, analyzing such data poses challenges as in how to combine and interpret the amount of information provided.
    \item Contaminated data: it refers to situations in which the data source is affected by noise from the environment, or in which there are missing values in the data, for instance as a result of different hardware or software malfunctions. The challenge arises from the fact that it is difficult to distinguish between true anomalies and different degrees of contamination. Furthermore, the contamination could strongly affect the anomaly detection.
    \item Big data: the challenge when working with big data comes from the overwhelming size of the data, as captured by the five V's: volume, velocity, variety, veracity and value.
\end{itemize}{}
\subsection{Anomalies in sensor systems}\label{anoalies_sensors}

Besides the aspects highlighted previously, the application domain and the specific scenario play an important role in choosing a technique to detect anomalies. Looking at sensor systems, common scenarios include Wireless Sensor Networks (WSN), Smart Cities, Smart Environments and, generally, IoT systems. Common sources of anomalies include the following \cite{Bosman_thesis}:
\begin{itemize}
    \item Environment: when the state of the environment changes, e.g. a new component is introduced, a highly unusual event occurs (such as natural disasters).
    \item System: the anomalies in this category are generated by the fact that a component of the system malfunctions or breaks down. The part of the system affected can be anything from a sensor to a cluster of nodes. The anomalies can be generated by hardware faults or limitations.
    \item Communication: these types of anomalies occur in systems which make use of different communication technologies (often wireless). They are caused by loss or delays of the communication packages. Part of these anomalies can be managed or avoided by making use of communication protocols in the applications.
    \item Various attacks: these anomalies are introduced in the system by a malevolent party. They are created by different means ranging from physical intervention on the system, to loading communication traffic.
\end{itemize}
Furthermore, besides the anomaly classification presented in \ref{anomalies_general}, for sensor systems we can also encounter the following categorization which is based on the faults noticed in real deployment \cite{Ni_sensor_fault}, \cite{Sharma_sensor_faults}, \cite{Yao_ad_sensor}:
\begin{itemize}
    \item Spike: a peak of a short duration in the recorded value. Usually, this distinctively deviates from the other measurements.
    \item Noise: there is an increase in the variance of a number of successive data samples. The real values can be strongly affected.
    \item Constant: a sensor reports as a measurement a constant value indifferent to the actual conditions. 
    \item Drift: an off-set is recorded in the measurements of the sensor values.
\end{itemize}
At this aim, authors in \cite{Ni_sensor_fault} present a more detailed taxonomy of sensor faults, with more categories, as well as possible causes and implications. In \cite{CAUTERUCCIO_ad_iot}, Cauteruccio et al. propose three different anomaly taxonomies, and formalize eight types of anomalies in the context of developing a framework for anomaly detection and classification within multiple IoT scenarios.

In sensor systems, most data to be analyzed is collected by the sensors connected to a processing unit. Classic sensors are more likely to collect data in the form of time series. Nevertheless, cameras also act as sensors, collecting images and videos. These are grouped under the category of spatiotemporal data. Anomalies are usually identified as unusual local changes in the spatial or temporal values. For spatial data (images), the surrounding neighbours of a data point play an important role in differentiating normal from abnormal data \cite{sun2004local}.


By looking at the research areas that have proposed various techniques for anomaly detection, we could characterize two main categories, namely: conventional techniques and data-driven techniques. In the first category we include techniques belonging to research fields with a long history in the topic, such as methods from  statistics and from signal processing. In the data-driven category we include techniques based on machine learning and data mining. These research fields  have become particularly active in the past few years, due to new technological advancements and current opportunities. In the following sections, we will present these two categories, with emphasis on the data-driven methods. It is important to note that these two categories are not mutually exclusive; there are in fact  solutions and algorithms that combine different techniques from both areas to address anomaly detection problems. Furthermore, there exist other categorizations of the methods used for outlier detection. For example, in \cite{AD_survey_Chandola} the authors group the methods as follows: classification based, clustering based, nearest neighbour based, statistical, information theoretic and spectral.

{
\subsection{Anomaly detection datasets}
When coping with anomaly detection problems, it is crucial to validate the effectiveness of a proposed technique onto real data. Indeed, since anomalies can occur in an unpredictable way and at any time, it can be difficult to generate them artificially. At this aim, industry and academia make available some datasets containing a number of anomalies ``fingerprints'' captured on field. Here we make a brief excursus about most credited anomaly detection datasets, which include both real and artificially generated data and cover multiple application domains.
}

{
The Bot-IoT dataset \cite{dataset_bot} contains a collection of normal and anomalous network traffic generated within a real IoT scenario at UNSW Canberra Cyber Lab. It is specifically exploited to validate techniques aimed at discriminating regular from anomalous flows within a sensors network. 
}

{
Outlier Detection DataSets (ODDS) \cite{ODDS} provides access to a collection of datasets (with ground truth if available) from different domains including: time series for event detection, crowded video scenes, opinion fraud detection data from online review systems.
}

{
The Numenta Anomaly Benchmark (NAB) \cite{lavin_evaluating_nab2015}, \cite{ahmad_unsupervised_NAB_2017} offers time series datasets concerned with real-time anomaly detection in streaming data generated by sensor systems. 
}

{
Yahoo Webscope \cite{laptev_generic_2015} includes a labeled anomaly detection dataset released by Yahoo Labs consisting of time series (both real and synthetic) with tagged anomaly points. 
}

{
UCI Machine Learning Repository \cite{uci_ml_repository:2019} comprises several hundred datasets (from videosurveillance to automotive domains) explicitly designed to evaluate machine learning techniques, including the ones allowing anomaly detection. 
}

{
Again, the ELKI outlier \cite{ELKI} dataset offers a collection of data whose details are available in a dedicated study \cite{campos_evaluation_datasets_2016}, where the authors focus on the analysis of unsupervised outlier detection algorithms.
}

{
Finally, the UCSD anomaly detection dataset \cite{ucsd} is a collection of images acquired by a camera overlooking pedestrian walkways; anomalies include the presence of bikers, skaters, small carts, and people walking across a walkway or in the grass that surrounds it. 
}

\section{Conventional techniques for anomaly detection} \label{conv_techniques}
This section provides an  overview of the most conventional anomaly detection techniques developed over the years. There is a long history of statistical methods that can be used to identify outliers. We should, however, note that even the most recent machine learning algorithms are grounded on their statistical counterparts \cite{Zimek_statistics_data_mining_ad_survey}. 
Furthermore, traditional probabilistic techniques are successfully being used by machine learning methods, e.g. Bayesian networks and Bayesian classifiers \cite{hruschka_bayesianML_2013}. In the following, we will consider a selection of conventional techniques, highlighting strengths and limitations, which leads to the more recent data-intensive approaches.

\subsection{Statistical methods}\label{stat_methods}

With statistical methods, it is assumed that the data points in a system are generated according to some statistical model. Any deviation from the expected model is regarded as an anomaly. In order to detect anomalies, statistic inference tests are applied. The types of techniques in this category can be parametric or non-parametric. For the parametric category, the underlying distribution of the data is known, and the parameters are estimated based on the data. Examples of parametric methods include those based on the Gaussian model, the Regression model, or a mixture of Parametric Distributions \cite{AD_survey_Chandola}. 

On the other hand, in the non-parametric case, the parameters of the underlying distribution are not known; they are determined from the existing data. Examples include methods based on histograms or on the Kernel Function. The work of \cite{hodge_outlier_detection_methodologies_2004} also considers the case of Proximity based techniques, such as k-NN, as well as that of Semi-Parametric methods, where local kernel models are applied instead of a single global distribution model. The research in \cite{markou_novelty_statistics_2003} provides an extended review of the types of statistical techniques used for novelty detection. It includes parametric approaches, such as probabilistic and Gaussian mixture modelling, Hidden Markov Models, Hypothesis testing, and non-parametric ones, such as k-NN based, Parzen density estimation, string matching, and clustering. Again, authors in \cite{pimentel_review_ND_2014} provide an updated review of novelty detection, including probabilistic approaches.

Statistical techniques have the advantage of being explainable as well as interpretable, especially when the distribution of the underlying data is known. In addition, some of these techniques, such as histogram-based or those which model single exponential distributions, can be easy to implement, or computationally efficient. However, kernel based techniques and models with complex distribution induce a higher computational complexity.

A limitation of the statistical approaches is that testing for outliers assumes that a specific distribution characterizes the recorded data points. This turns out not to be the case for most of the high dimensional data. Another challenge is that it may not be straightforward to choose the best statistics test for detecting the anomalies. Furthermore, histogram based techniques are not suited for multivariate data, as they do no take into account the interaction between the attributes of the data. When dealing with high dimensional data, machine learning techniques (Sect.\ref{data-driven-techniques}) tend to perform better,  as fewer assumptions are made about the actual data values.

\subsection{Time series analysis}\label{time_series}

It is often the case that the data in sensor systems is generated as a time series. Time series analysis (TSA) is concerned with methods that analyze the characteristics of the data, to extract useful statistics from them. Time series forecasting is often used to predict future expected values of the data to be recorded. The difference between the actual value and the expected one can be used for highlighting possible anomalies. 

Other common methods include cross-correlation analysis, auto regressive moving average (ARMA), auto regressive integrated moving average (ARIMA), Kalman filtering etc. Authors in \cite{mohamudally_ADE_2018} observe that the following time series models are often adopted in anomaly detection engines for the IoT: autoregressive models, symbolic TSA, seasonal trend loss decomposition, as well as combinations between machine learning techniques and TSA. In \cite{munir_fusead:_2019}, is advanced a proposal about fusing statistical models (ARIMA) and deep learning models (convolutional neural networks) for unsupervised anomaly detection. This approach takes advantage of the strengths of both techniques and outperforms the state-of-the-art of anomaly detection methods on a public dataset (Yahoo Webscope). The ablation and comparative study carried out by the authors shows that the fusion of the two techniques performs better than the individual components. 

Authors in \cite{serdio_fault_2014} suggest a residual-based approach for detecting faults in large-scale sensor networks by the use of vectorized auto-regressive moving average models (VARMA) and multivariate orthogonal space transformations. Three models are being proposed and compared against other existing techniques (state-of-the-art, linear, non-linear). One model (NFIR) together with the orthogonal transformations is likely to detect faults in the proposed problem. The authors' approach outperforms several state of the art solutions.

Time series analysis has the undoubted advantage of being simple and effective, since the pertinent outcomes can often be interpreted intuitively, especially when dealing with additive outliers. Unfortunately, it works well mainly for tracking "moderate" anomalous events. Unsatisfactory results are often achieved in cases when the anomalies are generated by more "dramatic" changes. 

\subsection{Signal processing}\label{sign_pro}
 
Signal processing deals with analyzing different types of signals, including sounds, images, outcomes of monitored physical processes, data streams from sensors, etc. Signal processing techniques are often adopted when signals are affected by noise, thus, specific de-noising mechanisms are put in place. This is needed to reveal possible anomalies also in presence of noise. 

Different transforms can be used to aid in anomaly detection such as the Fourier transform and the wavelet-based ones. For instance, \cite{rajagopalan_symbolicwavelet_2006} presents a wavelet-based approach for generating symbols for anomaly detection. The symbols are generated from the wavelet coefficients of the time series data. The authors show that the choice of an appropriate wavelet basis and scale greatly improves computational efficiency for real-time applications. The proposed technique is validated experimentally for a number of cases, as well as compared to another partitioning technique (symbolic false nearest neighbour). 

In \cite{brauckhoff_signal_2010}, authors analyze how temporal aggregation in random packet sampling alters the properties of the signal by introducing noise and aliasing that, in turn, affect the anomaly detection systems. They propose replacing the aggregation step with a specifically designed low-pass filter. As a result, aliasing is prevented and the new solution performs better in regards to the mis-detection and detection rates, even when considering a low sampling rate of the packets.
 
As occurs along most statistical methods, signal processing techniques exhibit a good detection rate when tackling the so-called "zero days" anomalies, namely, abnormal events that have never occurred before the current observation. By contrast, these techniques often rely on unrealistic assumptions (e.g. quasi-stationarity) of processes such as noise, which may detrimentally affect performance.

\subsection{Spectral techniques}\label{spectral} 
The spectral approach deals with dimensionality reduction. It is assumed that the data can be embedded into a lower dimensional subspace where the normal and anomalous instances greatly differ. Techniques can be based on Principal Component Analysis (PCA), a popular technique for projecting data into a lower dimensional space. 
Authors in \cite{huang_-network_nodate} propose a framework for network-wide anomaly detection based on distributed tracking and PCA. In \cite{egilmez_spectral_2014}, a proposal concerning a spectral anomaly detection method for WSNs by developing a graph-based filtering framework is advanced. The graphs are chosen as to include structural (proximity) information about the measured data. The authors show that standard PCA-based methods are a special case of their proposal. Their technique outperforms other state-of-the-art methods in global and distributed scenarios. 

The reduction of dimensionality characteristic of the spectral techniques represents an advantage when dealing with high dimensional data. Furthermore, it is suitable to be used as a preprocessing step before applying other types of anomaly detection techniques on the subspace. The drawbacks of spectral techniques often include high computational complexity, as well as not being applicable unless the anomalies can be separated from the normal conditions in a lower dimensional embedding of the data \cite{AD_survey_Chandola}.

\subsection{Information theory}\label{inf_theory} 

Methods in this category are aimed at analyzing the information content of a dataset by employing different information theoretic measures such as Kolmogorov complexity, entropy, relative entropy etc. The assumption behind these techniques is that anomalies distort the information content of the data instances in an otherwise normal dataset. Usually, the metrics are first computed using the whole dataset; then, a subset of points needs to be found. If this subset is eliminated, it induces the largest difference in value for the chosen metric. 
Authors in \cite{wenke_lee_information-theoretic_2001} propose several measures for anomaly detection, namely: entropy, conditional entropy, relative conditional entropy, information gain and information cost. These measures are used on three use cases to illustrate their utility. Furthermore, in \cite{ando_clustering_2007}, the information bottleneck formalization for outlier detection is exploited. 

The work in \cite{AD_survey_Chandola} outlines advantages and disadvantages of information theoretic methods. In particular, the specific information theoretic measure chosen for anomaly detection influences the performance of the various algorithms. It is often the case that anomalies are detected by the chosen measure only if there is a large number of anomalous data. Furthermore, when dealing with naturally ordered data, the information theoretic methods depend on determining an optimal substructure size to break the data. However, these techniques do not assume an underlying statistical distribution of the data, and can be used in an unsupervised fashion.

\subsection{General considerations about conventional techniques}

Conventional techniques for anomaly detection have been well assessed across the years, being also supported by robust and well established literature. In particular, these techniques allow for a rigorous quantification of an outlier, which may be detected through either of the following methods: 1) a deviation from an expected underlying model (statistical methods); 2) an abnormal metric value (e.g. the Euclidean distance) in measuring the difference between the expected and the predicted value (time series); 3) a frequency change highlighted in the Fourier transformed domain (signal processing); 4) a reconstruction error between the input and its projection on the eigenvectors space (PCA - spectral techniques); 5) a distortion in the  information content (Information theory).

All the aforementioned techniques mainly require a good knowledge of the so-called "ground truth", so as to derive a well quantifiable measure of the anomaly. Unfortunately, in many real-world scenarios where data models are extremely time-variant, the complexity of such techniques is not rewarded by satisfactory performance. This calls for the introduction of data-driven techniques which, at the expense of a less strict formalization, allow for more flexible adjustments, in line with the high dynamism of the Big Data paradigm.

\section{Data-driven techniques for anomaly detection}\label{data-driven-techniques}

Data-driven techniques typically refer to learning-based methods where the absence of a robust underlying mathematical model is compensated by the availability of large amounts of data, from which one can "learn" useful information.
Machine learning is a vast research field with many applications areas. Usually, it is classified into three main categories, namely supervised learning, unsupervised learning and reinforcement learning. Nevertheless, there also exist other combinations such as semi-supervised learning. Furthermore, with the technological advancements also deep learning is on the rise. Many machine learning techniques are often getting a deep approach or are combined with deep learning.

Similarly, Hodge and Austin \cite{hodge_outlier_detection_methodologies_2004} discern three fundamental approaches to the problem of outlier detection, namely:
\begin{itemize}
    \item Supervised: both the normality and abnormality are modeled; it requires labeled data for each of the categories;
    \item Unsupervised: identifying anomalies with no prior knowledge of the data;
    \item Semi-supervised: only normality is modeled; anomalies are identified by the fact that they are not within the normal threshold; it is also known as novelty detection or novelty recognition.

\end{itemize}
We can see that these three approaches present some overlap with the main categories of machine learning as they share common characteristics. In the next subsections we aim at providing the reader with a quick background of the main data-driven techniques, as well as discussing how these  are being used for anomaly detection in sensor systems. We look at the categories of machine learning techniques corresponding to the approaches outlined by Hodge and Austin \cite{hodge_outlier_detection_methodologies_2004}, adding new ones based on reinforcement learning and on deep learning.

\subsection{Supervised learning}
Supervised learning refers to machine learning techniques that train a model using a set of examples with a target output (labels). Supervised learning for anomaly detection has peculiar challenges when compared to other supervised learning applications. In practical cases, the rare samples are usually fewer in the training instances. 

There are modifications to algorithms for such class--imbalanced scenarios, to increase the impact of the rare instances onto the models. One popular approach is to  make use of cost-sensitive learning. The training data could be relabelled using the costs as it is the case with Metacost \cite{metacost}. In Metacost, the instances that have a reasonable probability of being in another class are relabelled to that class. In most cases it is the normal instances that get relabelled. 

Another type of cost-sensitive learning is the weighting method which applies a weight on the training instance that represents its misclassification cost. Such modifications have been made to common classification algorithms like the proximity-based classifiers \cite{mani2003knn}, Support Vector Machines (SVMs) \cite{aggarwal2012mining}, decision trees\cite{ting2002instance}, \cite{weiss2003learning}, and rule-based classifiers \cite{joshi2001mining}.

Supervised techniques exhibit great robustness since the "ground truth" is represented by pre-labeled data. Unfortunately, in many real systems, such an information is only partially available or not available at all. The introduction of semi-supervised and unsupervised methods fills this gap.

\subsection{Semi-supervised learning}

Semi--supervised learning generally refers to machine learning techniques that employ a small amount of labelled data and a large size of unlabelled data. Semi-supervised approaches can also be referred to machine learning techniques that train a classifier with 'normal' sensor data, such that the anomalies can be constructed and evolve dynamically \cite{javaid2016deep}. Though no explicit labels are used, this type of training dataset can be obtained in the real world through some form of labelling or separation of the normal data from a large pool of normal and abnormal data. 

Deep learning techniques like autoencoders, restricted Boltzmann \cite{fiore2013network}, have been applied for this learning task. In the case of an autoencoder, the model is trained with the normal data and it learns to reconstruct the input at a very small reconstruction error score. At detection, anomalies can be introduced. These anomalies would be reconstructed at a higher error score since they were not seen by the model during the training phase. A threshold is defined to capture the anomalous data. 

Chong and Tay have  applied this approach for the detection of anomalies in video data using, a spatio-temporal autoencoder \cite{chong2017abnormal}. The challenge with this technique, aside from the practical limitations of obtaining normal datasets, is that the optimal threshold can be challenging to define when obtained experimentally. One Class SVM (OC-SVM) is a type of semi-supervised SVM that does not require  labels of the anomaly. It was applied in \cite{garcia2016comparative} for the detection of attacks in smart city sensor networks.

Although the semi-supervised learning is the best choice to employ when only few labeled data are available, some limitations arise from assumptions in connection to the use of unlabeled data. These are based on relationships between labels and the unlabeled data distributions, whereby  bad assumptions may lead to poor performance.

\subsection{Unsupervised learning}
Unsupervised learning refers to learning from data where the  desired output is not available. A major challenge to deploying an anomaly detection system in sensor networks is the need for labelling data for use by learning algorithms. With unsupervised learning algorithms, we can build detection models without the need for manual labelling, thus reducing the deployment cost. Unsupervised learning algorithms are based on the assumption that the anomalies are rare and significantly different from the normal instances \cite{lee2001real}. 

One of the most popular methods is  based on clustering,  which uses a similarity measure to cluster data instances. Anomalies are identified as data instances that do not belong to clusters or which have clusters significantly smaller than the others. In \cite{rajasegarar2006distributed}, authors proposed a global outlier detection technique for anomaly detection in sensor nodes using clustering.

Other unsupervised learning algorithms are based on probabilistic modelling algorithms, whereby the anomalies are detected by estimating the likelihood of each data instance. An example is offered in the work of \cite{wang2008spatiotemporal}, a bayesian network was used as a form of unsupervised learning of the temporal and spatiotemporal data of a gas monitoring sensor network.

The supervised, semi-supervised, and unsupervised techniques work well in many data-driven situations. A general drawback of such methods is that they do not behave proactively when changes occur, unless an external guidance is provided. At this aim, more recent methods, such as reinforcement learning and deep learning, have been developed, both based on the possibility of learn autonomously.

\subsection{Reinforcement learning}\label{reinforcement_learning}
Sutton and Barto \cite{sutton_reinforcement_2018} explore a computation approach to learning from interaction, namely reinforcement learning (RL). This approach is the closest to how humans learn, and focuses on mapping situations to actions in order to maximize a goal through a numerical reward signal.

The main actors are a software agent, which can take actions, and the environment, whose state is affected by the actions that the agent takes. The action to be taken is chosen according to a policy which defines the behaviour of the agent. After each action, the agent receives a reward according to a reward function. The purpose of the agent is to maximize the total reward in the long run. 

Besides the reward function, which defines short-term desirability, an important element is the value function, which defines long-term desirability, namely the desirability of the possible states to follow.

What differentiates reinforcement learning from the other types of learning is its independence from supervision and focus on decision-making in pursuit of  a defined goal.

Reinforcement learning is often used in IoT scenarios with multi agent settings. It can be used in order to create a system that adapts to the environment. For example, Chincoli et al. have  proposed a protocol for controlling the transmission power by means of reinforcement learning in a multi agent system in IoT \cite{chincoli_self-learning_2018}. 

For reinforcement learning to be employed in the field of anomaly detection, more attention needs to be paid on how the problem can be formulated. Nevertheless, there exists some work in the area. Servin et al.   introduced distributed RL in a hierarchical architecture of network sensor agents \cite{servin_multi-agent_2008}. The system needs to interpret the signals and the states from a team of agents  to give an alarm in the case of an intrusion (intrusion detection). 
Oh et al. have looked  at anomaly detection for sequential data \cite{oh_sequential_2019}. The authors use Inverse Reinforcement Learning (IRL) to infer the reward function by looking at the sequence of actions taken by a target agent. In this way, outliers in trajectory data are identified. In \cite{huang_towards_2018} the authors suggest a time series anomaly detector based on a Recurrent Neural Network and RL. The RL method is used for the self-learning process.

While reinforcement learning is based on the concept of dynamically learn by tuning actions with the aim of maximizing a reward, novel deep-based approaches help to find patterns useful to make predictions on new data. Such operations are possible thanks to the presence of various (deep) layers,  characterizing the underlying artificial neural networks (ANN), so as to emulate a human brain.

\subsection{Deep learning}

Deep learning \cite{deepl-learning-book-goodfellow} emerged from the traditional ANNs. In terms of architecture, what makes deep learning different is that the hidden layers are more in depth (deep) than the traditional ANNs, which have  few hidden layers (shallow network). The additional layers enable deep learning to learn from massive data. 

Another difference is that,  in the more  conventional machine learning, the features are manually extracted from the input, before being fed into the network for the learning process. In deep learning, the features of the input are automatically learnt within the multiple layers. 

Deep learning can be used in either a supervised and unsupervised manner. Recent works have emerged that deployed deep learning for anomaly detection in sensor systems. Various types of deep learning networks have been proposed such as the Convolutional Neural Network (CNN), autoencoders, Restricted Boltzmann machine and the Recurrent Neural Network (RNN).

\textbf{Convolutional neural networks (CNNs)} work by exploiting three main concepts which are parameter sharing, sparse connectivity and equivariant representation \cite{sari2015review}. CNNs have been proven to be effective to extract features from samples and have been extensively applied to image, speech and text processing. The convolution layer is usually followed by the pooling and fully-connected layers for classification or regression tasks. CNNs can be used to autonomously learn useful features for an anomaly detection task \cite{janssens2016convolutional}. {Chen et al.  \cite{cnn_rev2_ad} propose the use of CNN for real-time anomaly detection in multi sensor signals (for the flash welding process). CNNs are used in order to learn the recurrence dynamics from the recurrence plots derived from the collected data. The proposed method was evaluated in both simulation studies and a real-world case study.}

\textbf{Autoencoders} are a special type of deep learning network that learns the latent space representations of data and tries to reconstruct the input data from the representations \cite{autoencoders-baldi}. The autoencoder aims to learn discriminative feature representations through a process of encoding and decoding. To enable the network to learn useful features, a bottleneck is introduced at the latent space to compress the data.
In \cite{chong2017abnormal} a spatiotemporal autoencoder is used to capture anomalies in camera networks.
Luo et al. \cite{luo2018distributed} proposed a distributed anomaly detection technique based on autoencoders to detect spikes and bursts recorded in temperature and humidity sensors. 

Even though autoencoders are an effective learning and detection technique, their performance can degenerate in the presence of noisy training data. Autoencoders can be used for anomaly detection when connected to a classification layer, as seen in the work by \cite{javaid2016deep}. Authors  applied a self-taught learning algorithm to solve anomaly detection for unforeseen and unpredictable scenarios in two stages. In the first stage, the features of an unlabeled data set are learnt by sparse autoencoders. The next step was feeding the features to a classifier like NB-Tree, Random Tree, or J48 trained on a labeled dataset. The labeled and unlabeled data must have relevance among them, even though they may come from different distributions.

\textbf{Sequential networks}, such as the Recurrent Neural Networks (RNNs), work by using temporal correlations between neurons \cite{RNN-Lipton15}. Recently, the Long-Short Term Memory (LSTM) was added to the RNN to serve as a memory unit during gradient descent \cite{sari2015review}. This addresses RNN's limited capacity to capture context with increase in the time steps. RNN and LSTM have been demonstrated to perform well in detecting anomalies in multivariate time series sensor data \cite{chalapathy2019deep}. Goh et al. \cite{goh2017anomaly} designed a time series predictor using RNN and used a Cumulative Sum method to identify the anomalies in a cyberphysical system. Their method was able to identify which sensor was attacked.
 
Chauhan et al. \cite{chauhan2015anomaly} proposed using LSTM to capture anomalies in Electrocardiography (ECG) signals. The LSTM was first used to design a predictive model of ECG signals. The probability distribution of the predicted errors was used to indicate whether a signal was normal or abnormal. One advantage of LSTM over other techniques is that data can be fed into the network without the need for pre-processing \cite{chauhan2015anomaly}.

\textbf{Generative adversarial networks (GANs)}, as introduced in \cite{goodfellow2014generative}, are among the most recent deep learning networks. A GAN works by training a generative network to generate fake samples, and then tries to fool the discriminator network until it can no longer differentiate between fake and real samples. 

GANs have been applied to identify anomalies in high dimensional and complex data from sensor systems in \cite{patel2018adversarial}, \cite{li2019mad} .
Patel et al. \cite{patel2018adversarial} introduced GANs for continuous real-time safety in learning-based control systems. They designed a controller focused anomaly detection in form an energy based GAN (EBGAN). The EBGAN network distinguishes between proper and anomalous actuator commands. However, when there are few anomalies, the more traditional machine learning techniques, like K-Nearest Neighbours, tend to perform better \cite{chalapathy2019deep}.

\textbf{Restricted Boltzmann machines (RBM)} are bipartite, fully-connected networks, having visible and hidden layers organized in an undirected graph \cite{RBM-introduction}. 
Authors in \cite{fiore2013network} applied RBM for network anomaly detection. 

When RBMs are stacked upon one another in multiple layers, the outcome is the \textit{Deep Belief Network} (DBN). The capabilities of DBN have been explored in intrusion detection system to detect attacks in the network \cite{alom2015intrusion}. 

DBNs can scale to large datasets and can improve interpretability as demonstrated by Wulsin et al. in \cite{wulsin2010semi}. Authors applied DBN for  electroencephalography (EEG) anomaly detection using it to separate rare and highly variable events, such as the malfunction of the brain. 

DBNs can be used in unsupervised learning as a pre-training method to train the parameters of the deep neural network;  after that, a supervised approach is used for classification.  This was the approach employed in \cite{kang2016intrusion} to detect anomalies and enhance security of in-vehicular networks.

\textbf{Hybrid models}. There have been efforts to combine different models to produce better results in anomaly detection. When a CNN serving as an encoder is combined with an LSTM network functioning as decoder, the result is a model effective for reconstructing image frames and detecting anomalies in data. Malhotra et al. \cite{malhotra2016lstm} combine LSTM and autoencoder for multi sensor anomaly detection. This works by reconstructing the normal behaviour of time series, and using the reconstruction errors to detect anomalies. The system captures anomalies that are caused by external factors or variables which are not captured by the numerous sensors monitoring a system. 

Zhou and Zhang \cite{zhou2015abnormal} combined sparse autoencoders and recurrent neural networks. The autoencoder was used for feature extraction, while the RNN was trained with a sequence of temporal features to predict the subsequent ones. 

Zhou et al. \cite{zhou2017anomaly} designed a denoising autoencoder using an anomaly regularizing penalty based on $L_1$ or $L_{2,1}$ norms to remove outliers from image datasets. 

Munawar et al. \cite{munawar2017spatio} combined CNN and RBM to extract features which when combined with a prediction system allow for learning to detect irregularities in video of industrial robot surveillance systems.

To sum up, deep learning techniques exhibit a high degree of adaptability but have a not negligible time complexity. This latter is due to the presence of many hidden layers (characterizing the deep approach) which require high training times. For this reason, within a sensor network deep learning cannot be performed directly on-board of a sensor, but requires nodes with dedicated computational resources typically located at the Edge, the Fog or in the Cloud (see Sect. \ref{sec:arch} for details about these architectures).  

\subsection{Online vs offline detection/algorithms}

Techniques such as deep learning pose the problem of "where" to process the huge amount of data. Another big issue is connected to the "when" such data is processed.  

There exists an increasing demand for accurate anomaly detection in streaming data and real-time systems. This is encouraged by the growth of the IoT and complex sensor based systems, which increase the availability of streaming data. The development of both hardware, software and architectural resources make the handling of streaming data and of strict requirements of real-time applications possible. 
A consequence of this situation is the shift from the more traditional offline processing and analysis of data to an online approach. Offline processing is concerned with analyzing the complete dataset, meaning that all the required data was collected and it is fully available. This approach usually allows for high--complexity techniques to be employed, as there are no time, nor computational constraints (assuming the computational part happens in powerful environments such as the Cloud). 

Contrarily, in online processing, there are limited computational resources, and time constraints for obtaining a result. Furthermore, the incoming data needs to be analyzed as it arrives, and any further processing happens in an online manner. Most techniques can be adapted to the online scenario by using short term memory (when the prediction is dependent only on a small number of previous measurements), windowing, regular updating of the model etc. \cite{DBLP:books/sp/Aggarwal2013}. 

Multiple examples of online sequential anomaly detection methods can be encountered in sensor systems and fields such as systems diagnosis and intrusion detection. \cite{Yao_ad_sensor} proposes an online anomaly detection algorithm for sensor data, namely segmented sequence analysis that leverages temporal and spatial correlations to build a piece wise linear model of the data. \cite{xu_stochastic_2017} introduces an online non parametric Bayesian method OLAD for detecting anomalies in streaming time series. \cite{ahmad_unsupervised_NAB_2017} uses Hierarchical Temporal Memory to develop an anomaly detection technique suited for unsupervised online anomaly detection.

The employment of online or offline algorithms must be done in accordance with the requirements of the system. The online approach is preferable when dealing with processing on the go on streaming data, real time or near real time requirements. Offline algorithms allow for carrying out more complex tasks on powerful resources when more data is available and when an immediate response is not necessary. The trade-offs between needed computational power, processing time, performance, response time and not only need to be taken into account when designing and employing these algorithms.

The techniques discussed till now highlighted pros and cons of various approaches by a methodological point of view. In fact, the effectiveness of a specific algorithm also depends on the architectural context of application. At this aim, in the next section we discuss the implications that three modern architectural models (Cloud, Fog, Edge) have onto sensor systems.

\section{Architectural perspective}
\label{sec:arch}

Sensor networks (and their ability of performing anomaly detection) can highly depend on the architectural model adopted to deploy them. We can distinguish the following models, as detailed next: the Cloud model; the Fog model; the Edge model; and hybrid combinations of the other ones.  

\subsection{Anomaly detection in the Cloud}\label{cloud} 

In the Cloud model \cite{cloud-computing-book-erl2013}, the information gathered by the sensors is processed in a virtually centralized environment, characterized by considerable computational resources. This would, for instance, be the case of a complex camera-based monitoring system, where the anomaly detection task results from a complex image-elaboration procedure \cite{Turchini_2018-dl-surveillance}.

Cloud-centred anomaly detection makes use of the virtually unlimited computational capabilities available in the Cloud, for the analysis of the data collected from sensor nodes. In this case, a major challenge derives from the actual "quality" of the data to be scrutinized. Sari et al. have looked at data security and storage issues, considering the analysis of sensor data that includes vast amounts of misconfigured sensor network traffic \cite{sari2015review}. 

Due to the inherent complexities and large-scale nature of virtual Clouds, the infrastructure itself is prone to software and hardware failures. Song et al. have looked at how to detect hardware and communication failures  \cite{fu2012hybrid}. They aimed at understanding complex, system--wide phenomena, to improve system and resource availability. 

There has been work on intrusion detection of networks in the Cloud environment, aimed at protecting sensor data and Cloud resources from malicious activities. Authors in \cite{pandeeswari2016anomaly} deployed an anomaly detector in a Cloud environment for network intrusion detection. The detector was trained with a hybrid algorithm, based on both Fuzzy C-Means clustering and Artificial Neural Networks (ANN). The Fuzzy C-Means was used to divide the large dataset into clusters to improve the learning capability of the neural network.

Despite being a reference method for anomaly detection, the Cloud-centric approach can be negatively affected by misconfigured traffic, due to the high volumes of incoming traffic \cite{sari2015review}. Cloud-centric anomaly detection may not be efficient for real-time applications due to the issues of latency, bandwidth and communication costs. As the number of sensors generating data increases, the issue of information bottleneck becomes more significant, since all raw data is required to reach the Cloud, before any processing may be applied. Delay become even more undesirable when a feedback control loop is involved between the Cloud and the physical devices, which is the case of wireless sensors and actuator networks \cite{akerberg_challenges-wsn-wsan-industrial_2011}.

The Cloud architectural model is well coupled with resource consuming techniques such as the deep-based ones. Obviously, such augmented "power" implies more costs, thus, a reasonable trade-off should be considered. A good strategy is to leave into Cloud only the critical functions, and going towards a "smoother" paradigm represented by Fog computing. 

\subsection{Anomaly detection in the Fog}\label{fog} 

In the Fog model \cite{mahmud_fog_2018},\cite{ZHANG_fog_fort} the information is processed on intermediate, Fog nodes lying between the cloud resources and the sensor system itself. This would, for instance be the case of a smart car environment, where an anomaly event (e.g. a generic self-driving malfunctioning) must be elaborated quickly, namely, as closely as possible to the source \cite{wang_real-time-ad-cav_2020}.

The need for introducing a model lying in the middle between a fully centralized and a fully decentralized approach was conceived alongside the Internet of Thing concept. In practice,  Fog computing can be considered as an extension of Cloud computing towards the sensors, with the aim of accelerating the information processing through the intermediate Fog nodes. 

Accordingly, some techniques have been devised to exploit the presence of such intermediate nodes, so as to also improve   anomaly detection. This is the case of \cite{FogSun}, where the authors aim at energy-efficiency by introducing the "virtual control nodes" to realize a cross-layer, clustering method directly on the sensing layer. 
A data reduction scheme is advanced in \cite{FogDeng}, where Fog nodes are able to build a prediction model fitting the sensor data characteristics, and resulting in a data stream reduction at the source node. 
Again, an energy optimization problem accounting for the presence of Fog nodes able to directly manipulate sensor measurements has been solved in \cite{FogFitzgerald}. 

Focused on evolved architectural schemes which exploit the power of the Fog paradigm are the works proposed in \cite{FogAwaisi} and \cite{FogWei}, where an efficient car parking architecture and a virtualization scheme for physical sensors are advanced, respectively. 

From an architectural viewpoint, Fog computing offers good chances to efficiently handle supervised techniques. For instance, an intensive training stage could be performed into Cloud, whereas the classification stage can occur on-board of sensors. This notwithstanding, in some situations (e.g. sensors located in areas with no data connection) the sensors might not have the access to the training set elaborated within the Cloud. In such cases, the sensors are compelled to perform each operation on their own, in line with the so-called Edge computing model.

\subsection{Anomaly detection at the Edge}\label{edge} 

In the Edge model \cite{yu_survey_edgeforIoT_2018}, the information can be directly processed on board of sensors, with options for distributed or collaborative decentralized computations. This would, for instance be the case of a WSN deployed to gather environmental parameters (e.g. temperature, CO$_2$, ground PH, etc.) \cite{borova_environmental_2019}. Often, such sensors are located quite far from a stable data connection; thus, the majority of information processing is executed on the sensor itself. 

Recently, several attempts have been made to develop machine learning algorithms that bring anomaly detection from the data centres closer to the sensor nodes. Software platforms like TensorFlow, Caffe, Tencent have toolboxes that enable lightweight, high-performance neural network anomaly detection processes that are suitable for Edge nodes. In \cite{schneible2017anomaly}, an Edge node equipped with a deep autoencoder model to detect anomalies is advanced. Authors in \cite{ALAM_edge_fort} propose a computation offloading model for the mobile edge with a focus on IoT applications. The model is based on deep reinforcement Q-learning and it shows improvements in terms of latency, energy consumption and execution time.

A range of lightweight machine learning methods (referred to as shallow-learning) have been adapted to run directly on sensor devices by Bosman et al. \cite{bosman_ensembles_2015}. They have demonstrated the viability of learning at the Edge on devices having as little as 20 kbytes of memory in non-floating point hardware. A range of anomalies may be detected directly in the sensor node, with improvements based on ensemble methods. 

Bosman et al. have also explored the intrinsic value of collaborative learning, involving data streams from neighboring sensors
\cite{bosman2017spatial}. Interestingly, they have shown that, by aggregating data from just a few sensor nodes, it is possible to substantially improve the accuracy of anomaly detection, still with no intervention from high-performance computing nodes. 

Ultimately, Edge solutions target the scalability issues arising from Cloud and (to some extent) Fog solutions, which incur network bottlenecks and feedback response latency. Nevertheless, the purely Edge-based solutions are limited in the type of anomalies that may be detected. More realistically, these are used as data pre-processing steps, and in combination with deeper learning methods that run in the Fog and in the Cloud, leading to the hybrid methods discussed next.

\subsection{Hybrid anomaly detection models}\label{hybrid} 

It is also possible to consider situations where different architectural models are used in combination, resulting in hybrid models. This is the case of a sensor network where part of the information processing (referred to as soft, or lightweight) is performed on board of the sensor (referred to as in-node anomaly detection), and the remaining critical part (hard processing) is performed either in the Fog, or in the Cloud, or in a mixed solution.

An exemplary case is offered by Cauteruccio et al. \cite{cauteruccio2019short}, who combined the strengths of Cloud and Edge nodes to detect anomalies in a heterogeneous sensor network. At the Edge nodes, an unsupervised neural network was deployed to  and detect short-term anomalies, or alerts. These are used to determine which portions of the data streams should be transmitted to the Cloud (along with the pre-processed alerts) for a computationally-intensive learning task. The authors show how this combination of short-term learning and long-term learning, based on a method called "multi-parametrized edit distance", may be used to achieve anomaly detection on different timescales. Furthermore, this hybrid (hierarchical) approach is shown to outperform other stand-alone strategies. 

Another exemplary case has been proposed by
Luo et al. \cite{luo2018distributed}. The system runs the more computationally-intensive tasks in the Cloud, and is able to offload specialized detection tasks to individual sensors, taking into account the capabilities of individual nodes. The whole framework is organized in a distributed computing fashion, which does not require synchronous coordination (or communication) among sensors and between sensors and Cloud. This offers some sort of asynchronous, hierarchical learning framework, which has rarely been explored in the literature. 

Still in the direction of distributed ML architectures, authors in \cite{zhang2019deep} have pinpointed important system-level issues that are not commonly addressed in centralized learning processes. These include: consistency, fault tolerance, communication, storage and resource management. 
Noteworthy frameworks for the execution of distributed ML processes, across different servers and sensor systems are MLbase \cite{kraska2013mlbase} and Gaia \cite{hsieh2017gaia}.

\begin{table*}[]
  \centering
  \caption{Promising Research Challenges}
    \begin{tabular}{ccl}
    \toprule
    \textbf{Challenge} & \textbf{Main Issue} & \multicolumn{1}{c}{\textbf{Research Topics}} \\
    \midrule
    \multirow{3}[2]{*}{Miniaturization} & \multirow{3}[2]{*}{ Hardware Limits} & $\cdot$ ML techniques for constrained devices \cite{bosman_ensembles_2015},\cite{bosman2017spatial} \\
          &       & $\cdot$ Distributed ML techniques \cite{kraska2013mlbase},\cite{hsieh2017gaia}\\
          &       & $\cdot$ Efficient fusion strategies \cite{bonawitz_federated_learning_2019}\\
    \midrule
    \multirow{2}[2]{*}{Acceleration} & \multirow{2}[2]{*}{ Circuitry Limits} & $\cdot$ FPGA-based optimization \cite{noauthor_accelerating_nodate},\cite{wess_neural_fpga_ecg_trade_2017} \\
          &       & $\cdot$ Algorithm complexity reduction \cite{wess_neural_fpga_ecg_trade_2017},\cite{mocanu_scalable_2018},\cite{sparse-dl-liu} \\
    \midrule
    \multirow{3}[2]{*}{Energy Efficiency} & \multirow{3}[2]{*}{Power Limits} & $\cdot$ Efficient routing protocols \cite{winter2012rpl}, \cite{rpl1},\cite{rpl2},\cite{fu_wsn_protocols} \\
          &       & $\cdot$ Smart topologies \cite{rpl1},\cite{rpl2} \\
          &       & $\cdot$ Lightweight operating systems \cite{tinyos}\\
    \midrule
    \multirow{3}[2]{*}{{Security}} & \multirow{3}[2]{*}{{Computational Limits}} & $\cdot$ {Distributed security mechanisms} \cite{sec_distr1},\cite{sec_distr2} \\
          &       & $\cdot$ {Novel keys distribution schemes} \cite{sec_key} \\
          &       & $\cdot$ {False data injection} \cite{sec_falseinj},\cite{sec_falseinj2}\\
    \midrule
     \multirow{2}[2]{*}{{Sensors Softwarization}} & \multirow{2}[2]{*}{{Special-purpose architectures}} & $\cdot$ {NFV/SDN for sensors} \cite{soft1}, \cite{soft2bis}, \cite{soft2}, \cite{hybrid-dl-ad-flow-detection-sdn}\\
          &       & $\cdot$ {Integration with the $5$G paradigm} \cite{soft3}, \cite{soft4} \\
    \midrule
    \multirow{2}[2]{*}{Architectural Models} & \multirow{2}[2]{*}{Intelligence Distribution} & $\cdot$ Hierarchical Learning \cite{pace_intelligence_edge_complexnetworks_2019}\\
          &       & $\cdot$ Hybrid models \cite{lavassani_fog_ml_iiot_2018},\cite{antonini_smart_audio_2018} \\
    \midrule
    \multirow{2}[2]{*}{Data Heterogeneity} & \multirow{2}[2]{*}{Aggregation} & $\cdot$ Ad-hoc middle-wares to  homogenize data \cite{Morshed2017DeepComputing}, \cite{casadei_datafusion}\\
          &       & $\cdot$ New fusion strategies accounting for data variability \cite{Morshed2017DeepComputing},\cite{han2012data}\\
    \bottomrule
    \end{tabular}%
  \label{tab:trend}%
\end{table*}%

\section{Open issues and challenges}\label{open-issues}
Anomaly detection in sensor systems brings about a broad range of possibilities, opening up to interesting "research" as well as "practical" realization challenges. Researchers will continue to be particularly intrigued by the promise of machine learning and hybrid software architectures, and the realization of hierarchical learning methods in a mixed Cloud-Fog-Edge setting. A pivotal issue is that, at its full extent, anomaly detection requires computational power, in neat contrast with the substantial hardware and software constraints of sensors. Next, we highlight promising trends and research challenges, as depicted in Table \ref{tab:trend}.

\subsection{Miniaturization}
Classic anomaly detection techniques are not designed to run in sensor systems, mainly due to their limited hardware capability, in terms of CPU, memory and connectivity. Some sensor nodes are even further constrained by their inability to compute floating-point operations, which Bosman et al. have demonstrated to cause  machine learning algorithms to become inaccurate and unstable \cite{bosman_ensembles_2015}. They have also provided practical guidelines as to how miniaturize and adapt a number of shallow-learning algorithms to stabilize. 

Limited accuracy arising from memory limitations have been addressed, to some extent, through ensemble learning \cite{bosman_ensembles_2015} and collaborative learning (involving data streams from neighboring sensors)
\cite{bosman2017spatial}. 
Yet, the miniaturization of machine learning algorithms to fit constrained devices remains a largely unsolved problem. This issue poses particularly hard hurdles when combined with the energy limitations of Edge devices, which are typically battery operated or even rely on fairly limited energy-harvesting techniques \cite{energy-harvesting-wsn}. 

Another promising research direction is "distributed machine learning",  where ML algorithms are re-designed to run in a multi-node environment. In this case, important issues include: parallel data management, fault tolerance, and communication/synchronization among nodes.  
Distributed machine learning systems like MLbase \cite{kraska2013mlbase} and Gaia \cite{hsieh2017gaia} have been developed to run ML algorithms across different servers and over large-scale sensor systems.
Moreover, authors in \cite{bonawitz_federated_learning_2019} proposed a system design for federated learning, which is a kind of distributed machine learning approach, to enable  model training using decentralized data from devices such as mobile phones. 

Both the distributed and the decentralized ML approaches present potential alternatives to centralized ML, particularly in the context of hybrid computing models. At the same time, new data-fusion strategies are required to consider the heterogeneity of data sources and handle the "missing data" problem \cite{missing-data-problem-ml}, which is particularly problematic in sensor systems, due to unreliable connectivity and sensor faults. 

\subsection{Acceleration}
Besides the algorithmic and software development in machine learning, there are also important developments in the hardware that may be used. This is mostly relevant to the server side of computation (in the Cloud and some Fog nodes) to perform the training of anomaly detection models on high performance computing platforms.

In the meantime, there is also progress in specialized acceleration hardware for Edge computing.
Several companies have invested, and recently released, chips and processors that can accelerate the execution of machine learning algorithms, often based on FPGA circuitry, to optimize some critical computations. For example, the Xilinx FPGA technology can be applied to the Deep Learning inference phase in order to achieve low latency, high compute performance, as well as low wattage \cite{noauthor_accelerating_nodate}. This opens the avenue for powerful techniques to be ran more closely to the sensor systems or even as part of them.

Again, Wess et al. \cite{wess_neural_fpga_ecg_trade_2017} have looked at how  ANNs may be effectively implemented on FPGA for anomaly detection, considering in EGC data. Their focus is on optimizing ANNs for FPGAs, based on  piece-wise linear approximated transfer functions and fixed point arithmetic. Furthermore, they carry out a resource trade-off analysis between the data point precision, size, latency and accuracy of the neural network, aimed at choosing a suitable ANN configuration (i.e. the number of input and hidden-layer neurons).

However, one must be aware that issues such as battery drainage and memory limitation will still exist in Edge computing. Besides acceleration based on hardware developments, considerable work is addressing the acceleration of the algorithms through advanced computing methods. Mocanu et al. 
have proposed an ingenious method to accelerate the ML training process, using network science strategies
\cite{mocanu_scalable_2018}. By training sparse rather than fully-connected ANNs, they achieved  a quadratic reduction in the number of parameters, at no accuracy loss. In fact, Liu et al. have demonstrated how deep learning over one million artificial neurons may be computed on commodity hardware, based on the sparse neural training strategy \cite{sparse-dl-liu}.

These developments suggest a promising research direction in ML acceleration based on bio-inspired algorithms, with enormous advantages arising from the possibility of making the existing algorithms more efficient, without changing the hardware platforms.

\subsection{Energy efficiency}
Some sensor systems such as WNSs have to deal with energy limitations, as these are often battery operated or rely on energy--harvesting. An important research strand is the investigation of the trade offs between algorithmic complexity and energy consumption, particularly in Edge devices. These energy constraints rule out the most sophisticated ML methods in Edge computing. Thus most investigations are directed towards shallow learning methods and simple reinforcement methods for in-node learning. An example is the work on shallow learning in sensors by Bosman et al. \cite{bosman_ensembles_2015}, \cite{bosman2017spatial}, as introduced in Sect. \ref{edge}. Also relevant is the work on Q-Learning (reinforcement learning) in sensors by Chincoli et al. \cite{chincoli_self-learning_2018}, as introduced in Sect. \ref{reinforcement_learning}. 

A common understanding so far is that the computational and energy overheads introduced by shallow learning is roughly counter-balanced by comparable savings arising from data compression and communication efficiency. Nevertheless, this is a very much debated issue, certainly worthy of more methodical investigation.  

Along this line of investigation, several researchers have focused on energy and communication efficiency that may be pursued through intelligent networks methods. The general issue is that communications incur significant energy overheads, often overshadowing the other neat energy-consumption contributors. What is more, in wireless networks, and particularly WSNs,  a significant portion of energy consumption derives from the sensing element, even while in standby mode \cite{KOTIAN-multihop}. Thus, WSN communication efficiency remains an open research challenge. 

A vast body of work has been looking at smart communication protocols, with new approaches to routing and topology control. A review of shallow-learning algorithms for energy efficiency based on intelligent power control methods has been written by Chincoli et al. \cite{Chincoli-tpc}. They have also demonstrated an autonomous self-learning method to minimize transmission power in WSNs, using in-node reinforcement learning \cite{chincoli_self-learning_2018}.

Assessing the impact that individual nodes have on the overall WSN consumption is a complex, non-deterministic problem, as discussed in \cite{KOTIAN-multihop}. To this end, cooperative learning offers a promising direction to address intelligent-based network efficiency. An example that is actually not employing ML but is using effective cooperative processes is offered by the Routing Protocol for Low-Power and Lossy Networks (or RPL) \cite{winter2012rpl}. RPL  is implemented in TinyOS \cite{tinyos}, i.e. a lightweight operating system designed to run in WSN-based environments. The core idea is for the protocol to create new paths and/or topologies, aiming at preserving the residual power of the various sensors.

Remarkably, the authors in \cite{rpl1} propose an approach aimed at forbidding low energy nodes to be forwarder nodes when the residual energy is lower than a specific threshold. Again, an interesting extension of RPL to mobility nodes (useful, for instance, within the automotive environment) is advanced in \cite{rpl2} where, by exploiting additional fields of RPL control packets, it is possible to advice a mobile sensor node for maintaining the energy by reducing network overheads.

Authors in \cite{fu_wsn_protocols} propose a multi-path routing protocol for WSNs which takes into account environmental data in order to assure the high routing reliability in harsh environmental conditions. Furthermore, the protocol favors the choice of routes which allow for high energy efficiency and low delivery latency.

Despite the vast body of work on energy efficiency, the use of intelligent algorithms has been limited to few cases, and is yet to find space in the standards. Cognitive networks, including in-node learning and inter-node collaborative learning, still have enormous research potential, particularly in the context of self-managed energy efficiency in wireless mesh networks, sensor networks and mobile ad-hoc networks.  

\subsection{{Security}}
{
Due to their limited resources, sensors cannot afford the luxury of hosting sophisticated protection mechanisms onboard; thus, they represent a very attractive prey for cyber-criminals.
From a malicious perspective, compromising a sensor might have a double purpose. On one hand, an attacker can be interested in stealing data or manipulating signals managed by a sensor through the so-called denial of service attacks \cite{bio_inspired_rev2,eusipco_ddos}. It is the case of medical devices  (e.g. insulin pumps, pacemakers, etc.) that, if compromised, can provoke irreversible damages to the patients \cite{sec_health}. On the other hand, a sensor can be deceitfully exploited as a propagating vector of a cyber-threat, whose avalanche effect globally amplifies the final malign result \cite{sec_spread,eusipco_kendall}. This is the case of the recent botnet Mirai \cite{sec_mirai}, based on a infectious propagation mechanism aimed at poisoning  IP-cameras systems. Such security breaches unavoidably impact also the sensors anomaly detection functionalities. 
Along these lines, many security frameworks and procedures are being proposed to tackle the aforementioned issues. 
}

{For instance, distributed security mechanisms applying the blockchain concepts to the sensors systems world have been proposed in \cite{sec_distr1} and \cite{sec_distr2}. Precisely, in the former work, a blockchain-based learning framework exploits the ``collective intelligence'' of a set of connected autonomous vehicles (CAV) aimed at improving the collective learning operation, where two approaches emerge: a centralized one, where each CAV uploads the sensing data into the cloud so that the model training is performed in a central server; a distributed one, where, according to a federated learning strategy, training data are distributed among learners, and the model training is performed locally. The latter work, instead,  proposes a Safety-as-a-Service framework, where distributed security and confidentiality mechanisms relying on the blockchain principles are applied into the realm of Industrial IoT. 
}

{An efficient random key distribution scheme (SRKD) to counter the replication attacks is instead proposed in \cite{sec_key}. Being specifically conceived for wireless sensor systems, such a scheme exhibits storage and communication overhead lower than classic counterparts. 
}

{Another prominent problem in constrained environments concerns the presence of malicious sensors which may inject anomalous data to corrupt the estimates at the fusion center (the so-called false data-injection attack). Accordingly, the authors in \cite{sec_falseinj} propose a detector to reveal time varying injection attacks within cyber-physical systems. The same family of attacks has been faced in \cite{sec_falseinj2}, where the authors introduce a method to reveal false data injection attacks against a hydraulic sensor-based system by exploiting the autoencoders. 
}

{
On another note, when looking at security within sensor systems and the IoT paradigm, one needs to also consider the in-network anomaly detection. Within the systems themselves, potential security risk needs to be assessed and identified based on network traffic data and behaviour. For example, the work in \cite{en-abc} investigates the increase of security in IoT-enabled application. The authors introduce a real-time multi-stage anomaly detection scheme which tries to cover the gaps with the traditional Density-Based Spatial Clustering of Applications with Noise (DBSCAN) algorithm. In \cite{multi-stage-ad-security-iot} an ensemble based anomaly detection technique is proposed to identify the malicious behaviour of nodes in the cloud environment. RBM and Unscented Kalman Filter are used for feature selection and optimization. The Artificial Bee Colony-based Fuzzy C-means is afterwards used to partition the datasets into relevant clusters. These clusters are then used to build a normal/abnormal profile of behaviour and to classify the occurring events as normal or anomalous.
}

{
At present, many efforts are being devoted to implement security mechanisms directly on board of a sensor, so that each device could react to cyber threats with no need of being supervised by a central server. Unluckily, due to the current limited technology, such approach seems still far from being used in real scenarios, but it would be supposed to replace the classic security-centric approach in the next few years, so as to make the sensors ever more independent and autonomous from a centralized architecture.
}

\subsection{{Sensors Softwarization}}
{
The emerging NFV (Network Function Virtualization) and SDN (Software Defined Networks) paradigms play a leading role within the process of network ``softwarization'', especially in modern $5$G infrastructures. NFV allows to decouple the underlying physical infrastructure (e.g. CPU, power supplies, etc.) of a network element from its software logic, whereas SDN allows to redefine the control plane of a network by enriching the routing and forwarding strategies. The combination of the two paradigms goes into the direction of improving flexibility and maintainability, and allows an effective cost sharing. 
Although the softwarization process mainly involves classic network nodes (e.g. routers, firewalls, etc.), a recent interest in adopting it within the sensor-based environments has emerged.      
}

{For instance, the authors in \cite{soft1} introduce a virtualized framework simulating a real IoT deployement, where virtual IoT honeynets are used to distract possible intruders from the real targets. The key idea is to transform the physical model into a common interoperable data model and, in turn, translate it into a software-based setting composed of Virtualized Network Functions (VNFs). The overall process is governed by NFV/SDN security policies, designed to dynamically handle operations such as filtering, dropping, and diverting.   
}

{
A combination of NFV and SDN is investigated in \cite{soft2bis} to automate network processes within an Internet-of-Vehicles (IoV) ecosystem. More specifically, NFV is exploited to offload IoV tasks towards the edge and cloud nodes, whereas SDN is utilized to smartly configure routing and forwarding paths across the IoV network.
}

{
The work in \cite{soft2} proposes an SDN-based framework to be exploited in IoT environments. In particular, two intelligent SDN controllers are advanced: the first one has the capability of estimating the packet flows within a specific sensing area through a partial recurrent spike neural network; and the second one exploits an ANN to select the cluster head and its members in the considered sensing area.  
}

{
Garg et al. \cite{hybrid-dl-ad-flow-detection-sdn} bring forward an SDN--based real-time anomaly detection framework for social multimedia traffic. The framework consists of two components, namely the anomaly detection module which leverages the advantages of improved RBM and SVM, and the end-to-end data delivery module of social media traffic. The latter comprises  an SDN-assisted multi-objective flow routing scheme, which allows for the trade-off between latency, bandwidth, and energy consumption utilization.
}

{
As key enablers of $5$G network infrastrucures, NFV/SDN concepts have been profitably exploited in \cite{soft3}, where a softwarized $5$G network has been designed to support the implementation of the so-called tactile internet, and to provide mission-critical IoT services.  
}
 
{
Again, the work in \cite{soft4} focuses on a $5$G-IoT architecture based on the multi-access edge computing (MEC) paradigm. In particular, the authors discuss the VNF life-cycle management through ad-hoc scheduling and allocation strategies allowing to dynamically instantiate, scale, migrate, and destroy the VNFs. 
} 

{Actually, sensors softwarization still has a long way to go, since virtualization technologies have been explicitly conceived for general-purpose architectures, whereas sensors often rely on very customized hardware (e.g. FPGA), which requires an additional effort to be managed.
}

\subsection{Architectural models}

When designing sensors systems, pinpointing a set of anomaly detection methods that can best fit the underlying architectural deployment is a non-trivial task. 
An important challenge is to identify techniques that are flexible enough as to be able to adjust to architectural changes -- for instance, from Cloud to Fog or  {\em vice versa}. It is equally crucial to dynamically manage the resources in a way that best employs the underlying infrastructure. An  example would be to automatically transfer the most demanding operations to the core network, when operating over the Cloud. Task offloading has been investigated extensively in the literature \cite{fernando_mobile_cloud_computing_2013}. Yet, its application to IoT and Cloud-assisted sensing still poses considerable open issues, given the variety and heterogeneity of frameworks and the significant constraints of Edge devices.

Accordingly, techniques such as hierarchical and distributed learning, which rely on a {\em divide et impera} approach, could be attractive when used in tandem with the novel paradigms of Cloud, Fog and Edge computing. A prominent area of investigation is aimed at better understanding how, and to which extent, it is possible to utilize the computing cycles available at the Edge to reduce network latency and bottlenecks, while at the same time respecting the energy limitations of Edge devices.

An example where machine learning may be used in a mixed architectural fashion is given by  authors in \cite{pace_intelligence_edge_complexnetworks_2019}. They illustrate the benefits of a hierarchical system versus a centralized one, corroborated by a use case involving cognitive transmission power control.

Distributed learning is explored in \cite{lavassani_fog_ml_iiot_2018}, where the authors experiment with a Fog architecture in an Industrial IoT context. 
The sensor motes learn a model from the incoming data, and periodically transmit updated parameters of the model to the Fog layer. This information is then sent to the Cloud where the data can be further analyzed and visualized. The results show high accuracy for simulating the original data, while minimizing the number of packets sent over the wireless link and the energy consumption. 

Likewise, the authors in \cite{antonini_smart_audio_2018} introduce a design framework for smart audio sensors which can record and pre-process the raw audio streams. The extracted features are transmitted to the Edge layer where anomaly detection algorithms executed as micro-services are able to detect anomalies in real-time by analyzing the received features.

These are just a few examples, showing the potential and the difficulties involved in the use of mix architectural models, under constrained hardware platforms. 

\begin{table*}
    \caption{Key references about techniques and architectural models}
    \label{tab:ref-table}
    \begin{tabular}{@{}lllll@{}}
    \toprule
    \multicolumn{5}{c}{\textbf{Conventional techniques}}                                                       \\ 
    \midrule
    Time series analysis           & \multicolumn{1}{l|}{\cite{mohamudally_ADE_2018},\cite{munir_fusead:_2019},\cite{serdio_fault_2014}}                  & \multirow{2}{*}{Statistical} & Non-parametric   & \cite{AD_survey_Chandola},\cite{hodge_outlier_detection_methodologies_2004},\cite{markou_novelty_statistics_2003},\cite{pimentel_review_ND_2014}   \\
    Signal processing              & \multicolumn{1}{l|}{\cite{rajagopalan_symbolicwavelet_2006},\cite{brauckhoff_signal_2010}}                   &                              & Parametric       & \cite{AD_survey_Chandola},\cite{hodge_outlier_detection_methodologies_2004},\cite{markou_novelty_statistics_2003},\cite{pimentel_review_ND_2014}   \\
    Information theory             & \multicolumn{1}{l|}{\cite{wenke_lee_information-theoretic_2001},\cite{ando_clustering_2007},\cite{AD_survey_Chandola}}                  & Spectral                     & \multicolumn{2}{l}{\cite{huang_-network_nodate},\cite{egilmez_spectral_2014},\cite{AD_survey_Chandola}} \\ 
    \midrule
    \multicolumn{5}{c}{\textbf{Data-driven techniques}}                                                        \\ \midrule
    Supervised                     & \multicolumn{1}{l|}{\cite{metacost},\cite{mani2003knn},\cite{aggarwal2012mining},\cite{ting2002instance},\cite{weiss2003learning},\cite{joshi2001mining}}                  & \multirow{6}{*}{Deep}        & RBM              & \cite{RBM-introduction},\cite{fiore2013network},\cite{alom2015intrusion},\cite{wulsin2010semi},\cite{kang2016intrusion}  \\
    Semi-supervised                & \multicolumn{1}{l|}{\cite{javaid2016deep},\cite{fiore2013network},\cite{chong2017abnormal},\cite{garcia2016comparative}}                    &                              & CNN              & \cite{sari2015review},\cite{janssens2016convolutional}  \\
    Unsupervised                   & \multicolumn{1}{l|}{\cite{lee2001real},\cite{rajasegarar2006distributed},\cite{wang2008spatiotemporal}}                   &                              & GAN              & \cite{goodfellow2014generative},\cite{patel2018adversarial},\cite{li2019mad},\cite{chalapathy2019deep}  \\
    \multirow{3}{*}{Reinforcement} & \multicolumn{1}{l|}{\multirow{3}{*}{\cite{sutton_reinforcement_2018},\cite{chincoli_self-learning_2018},\cite{servin_multi-agent_2008},\cite{oh_sequential_2019},\cite{huang_towards_2018}}} &                              & Autoencoder      & \cite{autoencoders-baldi},\cite{chong2017abnormal},\cite{luo2018distributed},\cite{javaid2016deep}  \\
                               &  \multicolumn{1}{l|}{}                  &                              & Sequential Nets.              & \cite{RNN-Lipton15},\cite{sari2015review},\cite{chalapathy2019deep},\cite{goh2017anomaly},\cite{chauhan2015anomaly}  \\ 
                            &  \multicolumn{1}{l|}{}                  &      & Hybrid models              & \cite{malhotra2016lstm},\cite{zhou2015abnormal},\cite{zhou2017anomaly},\cite{munawar2017spatio}  \\ 
    \midrule
    \multicolumn{5}{c}{\textbf{Architectural perspective}}                                                     \\ 
    \midrule
    Cloud                          & \multicolumn{4}{l}{\cite{cloud-computing-book-erl2013},\cite{sari2015review},\cite{fu2012hybrid},\cite{pandeeswari2016anomaly},\cite{Turchini_2018-dl-surveillance}}                                                     \\
    Fog                            & \multicolumn{4}{l}{\cite{mahmud_fog_2018},\cite{ZHANG_fog_fort},\cite{FogSun},\cite{FogDeng},\cite{FogFitzgerald},\cite{FogAwaisi},\cite{FogWei}}                                                     \\
    Edge                           & \multicolumn{4}{l}{\cite{yu_survey_edgeforIoT_2018},\cite{borova_environmental_2019},\cite{schneible2017anomaly},\cite{ALAM_edge_fort},\cite{bosman_ensembles_2015},\cite{bosman2017spatial}}                                                     \\
    Hybrid                         & \multicolumn{4}{l}{\cite{cauteruccio2019short},\cite{luo2018distributed},\cite{zhang2019deep},\cite{kraska2013mlbase},\cite{hsieh2017gaia}}                                                     \\ 
    \bottomrule
    \end{tabular}
\end{table*}

\subsection{Data Heterogeneity}
Although not directly connected to a specific anomaly detection technique or a particular architectural model, dealing with the data heterogeneity can be a very challenging task across the sensor systems world. This issue typically emerges when one is interested in extracting a particular ``meaning'' from a bulk of data having different shapes (e.g. temperature measurements and movement alerts), or coming from different sensor ecosystems (e.g. self driving cars and smart cameras). 

This is typically the realm of data fusion \cite{meng_survey_ml_data_fusion_2020}, which has recently turned its attention to smart methods in IoT settings \cite{ding_survey_data_fusion_iot_2019}, opening an avenue of new research issues. An important problem is the identification of anomalies that are not visible in individual (homogeneous) sensor data streams, and become evident only when heterogeneous streams (typically from neighboring sensors) are combined. Particularly difficult is the fusion of streams that operate on different time scales, different ranges and have different patterns.

Valuable examples are provided by healthcare applications, where data are generated from multiple sensors and sources, ranging from ECG reports (time series data), to blood test report (key value pairs), to x-rays (images). How to fuse these data into a single patient record to detect anomalies in the patient health record can be challenging.

Accordingly, there is a growing interest in using hybrid machine learning models to tackle heterogeneous sources \cite{Morshed2017DeepComputing}. The local models can be employed to learn specific complexities of sensor data and the models are aggregated to a global model in Edge and Cloud environments \cite{casadei_datafusion}. Similarly, several types of complex data are emerging with increasing heterogeneous sensors and applications. Anomalies will be detected in such complex data as time-series, sequential patterns, and biological sequences; graphs and networks; spatiotemporal data, including geospatial data, moving-object data, cyber-physical system data, and multimedia data \cite{han2012data}. 

\section{Conclusion}
In this review, we are focused on capturing the state-of-the-art of the anomaly detection problem across the sensor systems world. We have explored this timely problem along two main directions, as depicted in Fig.\ref{fig:feedback_impl}. First, we looked at two broad sets of techniques. The more conventional ones typically  exhibit robust mathematical formalism but are not always able to fully capture the complexity of real-world systems in deterministic models. Then, our emphasis shifted toward data-driven techniques that, by relying on machine learning concepts, aim at overcoming the non-linearity of sensor systems. In this case, however, the system is treated as a black box. Thus, the more realistic view achieved on the data behaviour sacrifices the formalism robustness, and creates the issues of data explainability and interpretability.   

Going beyond the individual anomaly detection methods, we explored the orthogonal strand of the architectural models that may be adopted for detecting anomalies in sensor systems. The most commonly ones include: 1) Cloud-assisted sensing, that follows the (virtually) centralized computing paradigm; 2) Fog in sensor systems, that follows the partially centralized paradigm; and 3) Edge sensing, the fully decentralized paradigm. 

As it turns out, the reality is often a mix \& match of different techniques (algorithms), models (deterministic vs predictive), architectures (Cloud, Fog, Edge), which makes our taxonomy very relevant to those who approach the complex area of anomaly detection in sensor systems. 
To facilitate the navigation of the most relevant works, we have clustered key papers in Table \ref{tab:ref-table}, which can be read in combination with the taxonomy introduced in Fig.\ref{fig:feedback_impl}

Our review sparks an interesting set of open research questions, as discussed in Sect. \ref{open-issues}, spanning from the miniaturization of the actual algorithms (to fit in constrained devices) to the acceleration of the processes (to scale up data analysis). We also discuss the issues of energy efficiency (particularly sensitive in devices), architectural models (to decentralize processes), and data heterogeneity (to improve accuracy by fusing data). Again, key references in relation to research challenges are summarized in Table \ref{tab:trend}. 

Our review recognized the important role that machine learning is playing in anomaly detection in sensor systems, identifying a range of important challenges that go beyond the development of suitable algorithms and lay at the intersection between computing (learning models), communications (efficiency) and engineering (constraints).

\vspace{-2mm}
\printcredits

\footnotesize

\bibliographystyle{IEEEtran} 
\bibliography{refs}

\vskip12pt

\makeatletter

\def\pct{\expandafter\@gobble\string\%}

\immediate\write\@auxout{\pct\space This is a test line.\pct }

\end{document}